\documentclass[sigconf,nonacm]{acmart}
\AtBeginDocument{}
\setcopyright{none}
\acmDOI{}
\acmISBN{}
\acmConference{}{}{}
\usepackage{booktabs}
\usepackage{array}
\usepackage{xspace}
\newif\ifpending
\pendingfalse
\newcommand{\T}{\textit{T}\xspace}
\newcommand{\ttau}{$\tau$\xspace}
\newcommand{\ci}[2]{[#1,\,#2]}
\newcommand{\atlas}{RubricAtlas\xspace}

\makeatletter
\renewcommand{\@mkauthors@iii}{%
  \global\setbox\mktitle@bx=\vbox{%
    \unvbox\mktitle@bx
    \centering
    {\large\bfseries
      Delin Wang\textsuperscript{1,3}, Sinuo Wang\textsuperscript{1,4},
      Rui Sun\textsuperscript{1}, Beidi Luan\textsuperscript{1},\\
      Qiming Shi\textsuperscript{1,5}, Jing Li\textsuperscript{1},
      Daxin Jiang\textsuperscript{1}, Zuo Bai\textsuperscript{1,2,\ensuremath{\dagger}}\par}
    \vspace{6pt}
    {\normalsize \textsuperscript{1}StepFun\qquad
      \textsuperscript{2}FinStep\qquad
      \textsuperscript{3}Renmin University of China\\
      \textsuperscript{4}University of Adelaide\qquad
      \textsuperscript{5}Zhejiang University\par}
    \vspace{4pt}
    {\small \textsuperscript{\ensuremath{\dagger}}Corresponding author:
      \href{mailto:baizuo@stepfun.com}{baizuo@stepfun.com}\par}
    \bigskip
  }%
}
\makeatother

\begin{document}
\newcommand{\numTypes}{99}
\newcommand{\numQuestions}{921}
\newcommand{\numTargets}{2{,}784}
\newcommand{\numItems}{8{,}160}
\newcommand{\numForbidden}{2{,}520}
\newcommand{\numConsensus}{503}
\newcommand{\numDiff}{418}
\newcommand{\numInstitutions}{64}
\newcommand{\numReportsPerType}{30}
\newcommand{\poolMin}{10}
\newcommand{\poolMax}{60}
\newcommand{\numReportTypes}{eleven}
\newcommand{\numIndustries}{31}
\newcommand{\numCells}{331}
\newcommand{\numQTypes}{13}
\newcommand{\numCandTypes}{106}
\newcommand{\numDeferred}{3}
\newcommand{\numDroppedTypes}{4}
\newcommand{\supThreshold}{3}
\newcommand{\wCons}{10}
\newcommand{\wDiff}{20}
\newcommand{\supMedian}{14}
\newcommand{\supCons}{21}
\newcommand{\consReach}{68}
\newcommand{\diffReach}{28}
\newcommand{\instMedian}{9}
\newcommand{\instDiff}{6}
\newcommand{\pctThreeInst}{94}
\newcommand{\pctFiveInst}{79}
\newcommand{\maxInstShare}{25}
\newcommand{\diffGeThree}{96}
\newcommand{\covReport}{54}
\newcommand{\covSource}{57}
\newcommand{\covTypeMin}{14}
\newcommand{\covTypeMax}{83}
\newcommand{\covLOTO}{53}
\newcommand{\covHalfOut}{56}
\newcommand{\inflSource}{10.5}
\newcommand{\inflSourceCIlow}{7.1}
\newcommand{\inflSourceCIhigh}{12.1}
\newcommand{\inflPosTasks}{77}
\newcommand{\inflRandom}{1.2}
\newcommand{\inflRandomMax}{4.5}
\newcommand{\tauSource}{0.60}
\newcommand{\tauRandomSubset}{0.81}
\newcommand{\kOneCov}{51}
\newcommand{\kOneTau}{0.69}
\newcommand{\kOneTauP}{0.51}
\newcommand{\kOneTauRand}{0.78}
\newcommand{\kOneTop}{24.0}
\newcommand{\kOneTopRand}{7.0}
\newcommand{\kOneInfl}{4.6}
\newcommand{\kOneInflRand}{0.7}
\newcommand{\kThreeCov}{79}
\newcommand{\kThreeTau}{0.79}
\newcommand{\kThreeTauP}{0.64}
\newcommand{\kThreeTauRand}{0.89}
\newcommand{\kThreeInfl}{2.0}
\newcommand{\kThreeInflRand}{0.3}
\newcommand{\kTenCov}{96}
\newcommand{\kTenTau}{0.95}
\newcommand{\kTenTauP}{0.87}
\newcommand{\kTenTauRand}{0.97}
\newcommand{\kTenInfl}{0.3}
\newcommand{\kTenInflRand}{0.0}
\newcommand{\smallPoolTau}{0.59}
\newcommand{\smallPoolCov}{45}
\newcommand{\midPoolTau}{0.79}
\newcommand{\numDraws}{200}
\newcommand{\numBoot}{2{,}000}
\newcommand{\tauTaskResample}{0.79}
\newcommand{\tauHalf}{0.84}
\newcommand{\tauHalfThree}{0.76}
\newcommand{\dropSourceQ}{14}
\newcommand{\dropSourceQTotal}{912}
\newcommand{\tauDropSource}{0.96}
\newcommand{\maxScoreChangeDropSource}{0.27}
\newcommand{\ruleJacMed}{0.10}
\newcommand{\ruleKappaMed}{0.16}
\newcommand{\ruleKappaMax}{0.29}
\newcommand{\ruleHitsDS}{11{,}956}
\newcommand{\ruleHitsLuna}{6{,}719}
\newcommand{\matchedPairs}{1{,}607}
\newcommand{\matchedDoubao}{54}
\newcommand{\matchedGLM}{60}
\newcommand{\matchedClaude}{52}
\newcommand{\searchesClaude}{53}
\newcommand{\fetchesClaude}{50}
\newcommand{\searchesGemini}{13}
\newcommand{\fetchesGemini}{5}
\newcommand{\corrOfficialT}{0.07}
\newcommand{\shPortal}{29}
\newcommand{\shOfficial}{20}
\newcommand{\numFetches}{15{,}798}
\newcommand{\numTraj}{980}
\newcommand{\leakPageShare}{1.7}
\newcommand{\leakReportsAny}{17}
\newcommand{\leakSrcN}{53}
\newcommand{\leakSourceTitle}{5.4}
\newcommand{\tauWeightOneOne}{0.96}
\newcommand{\tauWeightTwoOne}{0.73}
\newcommand{\tauWeightInvFreq}{0.91}
\newcommand{\tauWeightFreq}{0.78}
\newcommand{\tauConsOnly}{0.60}
\newcommand{\tauDiffOnly}{0.87}
\newcommand{\rhOneA}{1{,}073}
\newcommand{\rhOneB}{317}
\newcommand{\rhOneBoth}{114}
\newcommand{\rhOneKap}{0.12}
\newcommand{\rhTwoA}{2{,}481}
\newcommand{\rhTwoB}{1{,}896}
\newcommand{\rhTwoBoth}{875}
\newcommand{\rhTwoKap}{0.22}
\newcommand{\rhThreeA}{640}
\newcommand{\rhThreeB}{339}
\newcommand{\rhThreeBoth}{101}
\newcommand{\rhThreeKap}{0.17}
\newcommand{\rhFourA}{3{,}030}
\newcommand{\rhFourB}{1{,}194}
\newcommand{\rhFourBoth}{898}
\newcommand{\rhFourKap}{0.29}
\newcommand{\rhFiveA}{851}
\newcommand{\rhFiveB}{638}
\newcommand{\rhFiveBoth}{125}
\newcommand{\rhFiveKap}{0.10}
\newcommand{\rhSixA}{1{,}516}
\newcommand{\rhSixB}{1{,}362}
\newcommand{\rhSixBoth}{567}
\newcommand{\rhSixKap}{0.28}
\newcommand{\rhSevenA}{1{,}138}
\newcommand{\rhSevenB}{711}
\newcommand{\rhSevenBoth}{271}
\newcommand{\rhSevenKap}{0.22}
\newcommand{\rhEightA}{706}
\newcommand{\rhEightB}{193}
\newcommand{\rhEightBoth}{41}
\newcommand{\rhEightKap}{0.06}
\newcommand{\rhNineA}{341}
\newcommand{\rhNineB}{25}
\newcommand{\rhNineBoth}{8}
\newcommand{\rhNineKap}{0.04}
\newcommand{\rhTenA}{180}
\newcommand{\rhTenB}{44}
\newcommand{\rhTenBoth}{18}
\newcommand{\rhTenKap}{0.15}
\newcommand{\smallPoolHi}{20}
\newcommand{\midPoolLo}{21}
\newcommand{\midPoolHi}{39}
\newcommand{\matchedRatio}{20}
\newcommand{\numTasks}{99}
\newcommand{\numCommon}{98}
\newcommand{\numReports}{989}
\newcommand{\maxTurns}{1{,}000}
\newcommand{\maxTokens}{128k}
\newcommand{\cutoffDate}{2026-10-09}
\newcommand{\exSupport}{14}
\newcommand{\exInst}{13}
\newcommand{\exPool}{21}
\newcommand{\exTargets}{3}
\newcommand{\exItems}{9}
\newcommand{\numReportsPool}{3{,}079}
\newcommand{\scClaude}{59.5}
\newcommand{\hwClaude}{4.3}
\newcommand{\rkloClaude}{1}
\newcommand{\rkhiClaude}{2}
\newcommand{\pfClaude}{92}
\newcommand{\luClaude}{50.6}
\newcommand{\lupfClaude}{84}
\newcommand{\ctClaude}{67.1}
\newcommand{\dtClaude}{54.2}
\newcommand{\scDeepSeek}{56.5}
\newcommand{\hwDeepSeek}{4.3}
\newcommand{\rkloDeepSeek}{2}
\newcommand{\rkhiDeepSeek}{7}
\newcommand{\pfDeepSeek}{1}
\newcommand{\luDeepSeek}{47.5}
\newcommand{\lupfDeepSeek}{0}
\newcommand{\ctDeepSeek}{66.1}
\newcommand{\dtDeepSeek}{50.4}
\newcommand{\scDoubao}{55.5}
\newcommand{\hwDoubao}{4.3}
\newcommand{\rkloDoubao}{3}
\newcommand{\rkhiDoubao}{7}
\newcommand{\pfDoubao}{0}
\newcommand{\luDoubao}{47.6}
\newcommand{\lupfDoubao}{0}
\newcommand{\ctDoubao}{67.3}
\newcommand{\dtDoubao}{47.8}
\newcommand{\scGemini}{49.9}
\newcommand{\hwGemini}{3.8}
\newcommand{\rkloGemini}{8}
\newcommand{\rkhiGemini}{10}
\newcommand{\pfGemini}{0}
\newcommand{\luGemini}{43.6}
\newcommand{\lupfGemini}{0}
\newcommand{\ctGemini}{59.4}
\newcommand{\dtGemini}{43.1}
\newcommand{\scGLM}{57.5}
\newcommand{\hwGLM}{4.1}
\newcommand{\rkloGLM}{1}
\newcommand{\rkhiGLM}{6}
\newcommand{\pfGLM}{5}
\newcommand{\luGLM}{49.6}
\newcommand{\lupfGLM}{15}
\newcommand{\ctGLM}{67.4}
\newcommand{\dtGLM}{51.1}
\newcommand{\scHunyuan}{56.0}
\newcommand{\hwHunyuan}{4.5}
\newcommand{\rkloHunyuan}{2}
\newcommand{\rkhiHunyuan}{7}
\newcommand{\pfHunyuan}{0}
\newcommand{\luHunyuan}{47.8}
\newcommand{\lupfHunyuan}{0}
\newcommand{\ctHunyuan}{64.6}
\newcommand{\dtHunyuan}{49.9}
\newcommand{\scKimi}{50.5}
\newcommand{\hwKimi}{4.3}
\newcommand{\rkloKimi}{8}
\newcommand{\rkhiKimi}{10}
\newcommand{\pfKimi}{0}
\newcommand{\luKimi}{43.9}
\newcommand{\lupfKimi}{0}
\newcommand{\ctKimi}{61.0}
\newcommand{\dtKimi}{43.0}
\newcommand{\scMimoFlash}{57.0}
\newcommand{\hwMimoFlash}{4.6}
\newcommand{\rkloMimoFlash}{2}
\newcommand{\rkhiMimoFlash}{7}
\newcommand{\pfMimoFlash}{2}
\newcommand{\luMimoFlash}{48.3}
\newcommand{\lupfMimoFlash}{1}
\newcommand{\ctMimoFlash}{65.5}
\newcommand{\dtMimoFlash}{51.8}
\newcommand{\scMimoPro}{57.0}
\newcommand{\hwMimoPro}{4.3}
\newcommand{\rkloMimoPro}{2}
\newcommand{\rkhiMimoPro}{7}
\newcommand{\pfMimoPro}{1}
\newcommand{\luMimoPro}{48.1}
\newcommand{\lupfMimoPro}{1}
\newcommand{\ctMimoPro}{65.8}
\newcommand{\dtMimoPro}{51.1}
\newcommand{\scMiniMax}{49.4}
\newcommand{\hwMiniMax}{4.1}
\newcommand{\rkloMiniMax}{8}
\newcommand{\rkhiMiniMax}{10}
\newcommand{\pfMiniMax}{0}
\newcommand{\luMiniMax}{42.2}
\newcommand{\lupfMiniMax}{0}
\newcommand{\ctMiniMax}{60.9}
\newcommand{\dtMiniMax}{41.7}
\newcommand{\marginA}{2.0}
\newcommand{\marginAlo}{$-$0.7}
\newcommand{\marginAhi}{4.7}
\newcommand{\marginB}{1.0}
\newcommand{\marginBlo}{$-$1.4}
\newcommand{\marginBhi}{3.4}
\newcommand{\spreadThreeSevenA}{1.5}
\newcommand{\spreadThreeSevenB}{0.8}
\newcommand{\gapSevenEightA}{5.0}
\newcommand{\gapSevenEightB}{3.6}
\newcommand{\tauAB}{0.87}
\newcommand{\levelGap}{8.0}
\newcommand{\pairsOne}{36}
\newcommand{\flipsOne}{1}
\newcommand{\pairsTwo}{28}
\newcommand{\numQCompared}{9{,}196}
\newcommand{\qCorrT}{0.83}
\newcommand{\numCkCompared}{81{,}469}
\newcommand{\ckAgree}{85.4}
\newcommand{\ckKappa}{0.71}
\newcommand{\ckJaccard}{0.75}
\newcommand{\ckYesA}{56}
\newcommand{\ckYesB}{48}
\newcommand{\numNcCompared}{25{,}164}
\newcommand{\ncAgree}{99.0}
\newcommand{\ncKappa}{0.39}
\newcommand{\ncTrigA}{262}
\newcommand{\ncTrigB}{151}
\newcommand{\consSDA}{2.8}
\newcommand{\diffSDA}{4.1}
\newcommand{\sdGapA}{1.3}
\newcommand{\sdGapLoA}{0.33}
\newcommand{\sdGapHiA}{2.3}
\newcommand{\diffLeadA}{2.4}
\newcommand{\diffFirstShareA}{86}
\newcommand{\consSDB}{2.1}
\newcommand{\diffSDB}{3.1}
\newcommand{\sdGapB}{0.9}
\newcommand{\sdGapLoB}{0.00}
\newcommand{\sdGapHiB}{1.9}
\newcommand{\diffLeadB}{0.8}
\newcommand{\diffFirstShareB}{55}
\newcommand{\consFirstShareGLMA}{39}
\newcommand{\consFirstShareDoubaoA}{34}
\newcommand{\consFirstShareClaudeA}{21}
\newcommand{\jnN}{100}
\newcommand{\jnTasks}{10}
\newcommand{\jnCorr}{0.96}
\newcommand{\jnSD}{4.1}
\newcommand{\jnModelSD}{0.42}
\newcommand{\jnShift}{1.8}
\newcommand{\jnLevel}{39.6}
\newcommand{\rbLowA}{0.94}
\newcommand{\rbHighA}{0.96}
\newcommand{\rbMedB}{0.93}
\newcommand{\rbFullB}{0.95}
\newcommand{\tokMedA}{8.1k}
\newcommand{\tokLowA}{5.8k}
\newcommand{\tokMedB}{1.6k}
\newcommand{\tokHighB}{3.5k}
\newcommand{\delNA}{27}
\newcommand{\delBeforeA}{0.64}
\newcommand{\delAfterA}{0.21}
\newcommand{\delKA}{20}
\newcommand{\delLoA}{55}
\newcommand{\delHiA}{87}
\newcommand{\delRelA}{71}
\newcommand{\delOtherA}{0.033}
\newcommand{\delOtherNoopA}{0.025}
\newcommand{\delShareAQ}{0.19}
\newcommand{\delShareNoopA}{0.11}
\newcommand{\padIrrA}{$-$1.4}
\newcommand{\padIrrLoA}{$-$3.4}
\newcommand{\padIrrHiA}{0.5}
\newcommand{\padIrrNetA}{$+$1.1}
\newcommand{\padDupA}{$-$2.1}
\newcommand{\padDupLoA}{$-$4.2}
\newcommand{\padDupHiA}{0.1}
\newcommand{\padDupNetA}{$+$0.5}
\newcommand{\noopA}{$-$2.5}
\newcommand{\delNB}{27}
\newcommand{\delBeforeB}{0.57}
\newcommand{\delAfterB}{0.19}
\newcommand{\delKB}{22}
\newcommand{\delLoB}{63}
\newcommand{\delHiB}{92}
\newcommand{\delRelB}{83}
\newcommand{\delOtherB}{0.028}
\newcommand{\delOtherNoopB}{0.029}
\newcommand{\delShareBQ}{0.19}
\newcommand{\delShareNoopB}{0.08}
\newcommand{\padIrrB}{$+$0.8}
\newcommand{\padIrrLoB}{$-$1.2}
\newcommand{\padIrrHiB}{2.9}
\newcommand{\padIrrNetB}{$+$0.8}
\newcommand{\padDupB}{0.0}
\newcommand{\padDupLoB}{$-$1.7}
\newcommand{\padDupHiB}{1.7}
\newcommand{\padDupNetB}{0.0}
\newcommand{\ptN}{30}
\newcommand{\ptPad}{40}
\newcommand{\rejB}{5}
\newcommand{\rejNoise}{20}
\newcommand{\rejPert}{15}
\newcommand{\corrSearchT}{0.26}
\newcommand{\corrFetchT}{0.24}
\newcommand{\corrSearchTB}{0.18}
\newcommand{\corrFetchTB}{0.16}
\newcommand{\corrFetchLen}{0.44}
\newcommand{\regFetchA}{1.2}
\newcommand{\regFetchASE}{0.55}
\newcommand{\regSearchA}{1.1}
\newcommand{\regSearchASE}{0.9}
\newcommand{\regFetchLenA}{1.0}
\newcommand{\regFetchLenASE}{0.54}
\newcommand{\regLenA}{8.6}
\newcommand{\regLenASE}{1.8}
\newcommand{\regFetchB}{0.3}
\newcommand{\regFetchBSE}{0.57}
\newcommand{\lenCoefA}{9.2}
\newcommand{\lenSEA}{1.7}
\newcommand{\lenCoefB}{5.7}
\newcommand{\lenSEB}{1.3}
\newcommand{\lenLeaderA}{Claude}
\newcommand{\shiftLenR}{0.82}
\newcommand{\shiftClaude}{$+$0.9}
\newcommand{\shiftloClaude}{$-$0.4}
\newcommand{\shifthiClaude}{2.2}
\newcommand{\shiftKimi}{$-$1.4}
\newcommand{\shiftloKimi}{$-$2.8}
\newcommand{\shifthiKimi}{0.0}
\newcommand{\shiftDeepSeek}{$+$1.0}
\newcommand{\shiftloDeepSeek}{$-$0.5}
\newcommand{\shifthiDeepSeek}{2.5}
\newcommand{\dsResidual}{$+$0.4}
\newcommand{\dsResidualLo}{$-$1.4}
\newcommand{\dsResidualHi}{2.1}
\newcommand{\leakSellShare}{31}
\newcommand{\leakSellDelta}{$+$0.5}
\newcommand{\tauWeightOneThree}{0.96}
\newcommand{\largePoolTau}{0.65}
\newcommand{\largePoolLo}{40}
\newcommand{\largePoolHi}{60}
\newcommand{\diffThreeInst}{90}
\newcommand{\flagTau}{0.53}
\newcommand{\flagR}{0.76}
\newcommand{\flagMinA}{62}
\newcommand{\flagMaxA}{75}
\newcommand{\flagMinB}{39}
\newcommand{\flagMaxB}{63}
\newcommand{\flagLeastA}{Doubao}
\newcommand{\flagDoubaoA}{61.5}
\newcommand{\flagDoubaoB}{38.6}
\newcommand{\mtOne}{0.18}
\newcommand{\mrOne}{0.57}
\newcommand{\mtTwo}{0.47}
\newcommand{\mrTwo}{0.64}
\newcommand{\mtThree}{0.69}
\newcommand{\mrThree}{0.73}
\newcommand{\mtFour}{0.60}
\newcommand{\mrFour}{0.75}
\newcommand{\mrateAFour}{32.9}
\newcommand{\mrateBFour}{13.0}
\newcommand{\mtFive}{0.49}
\newcommand{\mrFive}{0.66}
\newcommand{\mtSix}{0.22}
\newcommand{\mrSix}{0.49}
\newcommand{\mtSeven}{0.64}
\newcommand{\mrSeven}{0.91}
\newcommand{\mtEight}{0.69}
\newcommand{\mrEight}{0.76}
\newcommand{\mtNine}{$-$0.04}
\newcommand{\mrNine}{$-$0.12}
\newcommand{\mtTen}{0.09}
\newcommand{\mrTen}{0.46}
\newcommand{\mtMed}{0.48}
\newcommand{\padIrrNetLoA}{$-$0.9}
\newcommand{\padIrrNetHiA}{3.2}
\newcommand{\padDupNetLoA}{$-$1.3}
\newcommand{\padDupNetHiA}{2.5}
\newcommand{\padExpectA}{3.1}
\newcommand{\lcTauA}{0.73}
\newcommand{\lcSecondA}{MimoPro}
\newcommand{\padIrrNetLoB}{$-$1.4}
\newcommand{\padIrrNetHiB}{3.0}
\newcommand{\padDupNetLoB}{$-$2.1}
\newcommand{\padDupNetHiB}{2.1}
\newcommand{\lcTauB}{0.87}
\newcommand{\lcSecondB}{GLM}
\newcommand{\compInfl}{5.6}
\newcommand{\compInflMax}{8.6}
\newcommand{\compTau}{0.74}
\newcommand{\compTauP}{0.56}
\newcommand{\consShareSrc}{71}
\newcommand{\consShareAll}{55}
\newcommand{\inflBeyondComp}{4.9}
\newcommand{\covGenreHi}{73}
\newcommand{\covGenreLo}{33}
\newcommand{\covShort}{50}
\newcommand{\covLong}{63}
\newcommand{\tauABlo}{0.56}
\newcommand{\tauABhi}{0.96}
\newcommand{\flagTauLo}{0.33}
\newcommand{\flagTauHi}{0.78}
\newcommand{\ckKappaLo}{0.70}
\newcommand{\ckKappaHi}{0.72}
\newcommand{\pairsOneB}{34}
\newcommand{\leakSrcLo}{4.2}
\newcommand{\leakSrcHi}{7.0}
\newcommand{\leakSrcEff}{$-$0.7}
\newcommand{\leakSrcEffLo}{$-$3.5}
\newcommand{\leakSrcEffHi}{2.1}
\newcommand{\lurkloClaude}{1}
\newcommand{\lurkhiClaude}{3}
\newcommand{\luhwClaude}{4.1}
\newcommand{\lurkloDeepSeek}{3}
\newcommand{\lurkhiDeepSeek}{7}
\newcommand{\luhwDeepSeek}{4.2}
\newcommand{\lurkloDoubao}{2}
\newcommand{\lurkhiDoubao}{7}
\newcommand{\luhwDoubao}{4.0}
\newcommand{\lurkloGemini}{8}
\newcommand{\lurkhiGemini}{10}
\newcommand{\luhwGemini}{3.7}
\newcommand{\lurkloGLM}{1}
\newcommand{\lurkhiGLM}{4}
\newcommand{\luhwGLM}{3.9}
\newcommand{\lurkloHunyuan}{3}
\newcommand{\lurkhiHunyuan}{7}
\newcommand{\luhwHunyuan}{4.1}
\newcommand{\lurkloKimi}{8}
\newcommand{\lurkhiKimi}{10}
\newcommand{\luhwKimi}{4.0}
\newcommand{\lurkloMimoFlash}{2}
\newcommand{\lurkhiMimoFlash}{7}
\newcommand{\luhwMimoFlash}{4.2}
\newcommand{\lurkloMimoPro}{2}
\newcommand{\lurkhiMimoPro}{7}
\newcommand{\luhwMimoPro}{4.0}
\newcommand{\lurkloMiniMax}{8}
\newcommand{\lurkhiMiniMax}{10}
\newcommand{\luhwMiniMax}{3.7}
\newcommand{\tokHighAexact}{8{,}127}
\newcommand{\tokMedAexact}{8{,}137}
\newcommand{\mrateBEightPct}{2.1}
\newcommand{\srcTopThreeChanged}{76}
\newcommand{\srcTopOneChanged}{50}
\newcommand{\randTopOneChanged}{45}
\newcommand{\randTopOneLo}{38}
\newcommand{\randTopOneHi}{52}
\newcommand{\verbN}{800}
\newcommand{\verbSpans}{3{,}881}
\newcommand{\verbTexts}{3{,}228}
\newcommand{\repoURL}{TBD}

\title[RubricAtlas: Type-Level Rubrics for Deep Research]{RubricAtlas: Mining Type-Level Rubrics from Expert Reports to Evaluate Deep Research Agents}
\author{Delin Wang}
\affiliation{\institution{StepFun}\city{Shanghai}\country{China}}
\affiliation{\institution{Renmin University of China}\city{Beijing}\country{China}}
\author{Sinuo Wang}
\affiliation{\institution{StepFun}\country{China}}
\affiliation{\institution{University of Adelaide}\city{Adelaide}\country{Australia}}
\author{Rui Sun}
\affiliation{\institution{StepFun}\country{China}}
\author{Beidi Luan}
\affiliation{\institution{StepFun}\country{China}}
\author{Qiming Shi}
\affiliation{\institution{StepFun}\country{China}}
\affiliation{\institution{Zhejiang University}\city{Hangzhou}\country{China}}
\author{Jing Li}
\affiliation{\institution{StepFun}\country{China}}
\author{Daxin Jiang}
\affiliation{\institution{StepFun}\country{China}}
\author{Zuo Bai}
\email{baizuo@stepfun.com}
\affiliation{\institution{StepFun}\country{China}}
\affiliation{\institution{FinStep}}
\renewcommand{\shortauthors}{Wang et al.}

\begin{abstract}
Deep research agents conduct multi-step Web searches and synthesise retrieved information into analytical reports. Evaluating these outputs requires checking which analyses they deliver, beyond the presence of relevant facts. Query-specific binary rubrics make individual requirements explicit, but task-level checklists may leave the overall analytical structure underexamined. We introduce \atlas, which derives \emph{type-level analytical requirements} from expert reports on different objects within the same research type and organises detailed checks around recurring analytical questions. The requirements give related tasks a shared standard for what to analyse, without prescribing their answers. Our framework links binary checkpoints to report passages and evaluates completeness separately from argument validity. Built from \numReportsPool{} Chinese financial research reports, \atlas covers \numTypes{} research types. Reference-report selection experiments show that individual reports cover only about half of their type's analytical questions, and that selecting requirements from different reports changes evaluation scores and agent rankings. Across ten agents, two judge pipelines achieve checkpoint agreement of $\kappa$ = \ckKappa{} and agent-ranking agreement of Kendall $\tau$ = \tauAB{}. Controlled deletions reduce scores on the affected analyses, while padding produces no detectable increase. These findings support type-level rubrics for comparing the analytical completeness of Web deep research outputs, with individual decisions traceable to report evidence.
\end{abstract}

\maketitle

\section{Introduction}
Deep research agents carry out multi-step Web searches, read retrieved pages and synthesise findings into analytical reports~\cite{nakano2021webgpt,du2025drbench,gou2025mind2web2}. Unlike short-answer Web tasks~\cite{wei2025browsecomp,mialon2023gaia}, these reports often require more than factual retrieval, including comparisons, explanations and examination of the conditions behind their conclusions. A report on an oil and gas producer, for example, may list production, reserves and costs without comparing the firm with peers or explaining whether its advantage can persist. Including relevant facts is not the same as completing the relevant analysis.

A common approach to evaluating deep research outputs is to use \emph{query-specific binary rubrics}: requirements written for an individual task and judged one by one as met or unmet~\cite{sharma2025researchrubrics,akyurek2025prbench,li2026drbench2}. These rubrics make individual requirements explicit. However, an evaluation centred on checklist items can foreground task-specific details while leaving the report's overall analytical structure underexamined. The issue is not whether decisions are binary, but whether they reveal which analyses the details collectively support and which remain incomplete. We therefore organise evaluation around analytical questions, with detailed checks specifying what each analysis should deliver. To ground this organisation beyond the choices made for one query or reference report, we derive recurring requirements from expert reports on different research objects.

Expert reports provide examples of how professionals analyse different objects. We define a \emph{research type} as a research-question family within an industry, such as studies of oil and gas producers' operations and value realisation. Within a type, reports may attribute advantages to different resources or distribution networks, but repeatedly ask: \emph{which verifiable capabilities distinguish the firm from its peers, and under what conditions can that advantage persist?} By combining these reports, we identify recurring requirements: name the capability, compare it with named peers using comparable metrics, and examine what could sustain or erode the advantage. These requirements describe what to analyse, not what to conclude. They reflect the source reports' professional practice, rather than a universal checklist for every request.

\begin{figure*}[t]
\includegraphics[width=\textwidth]{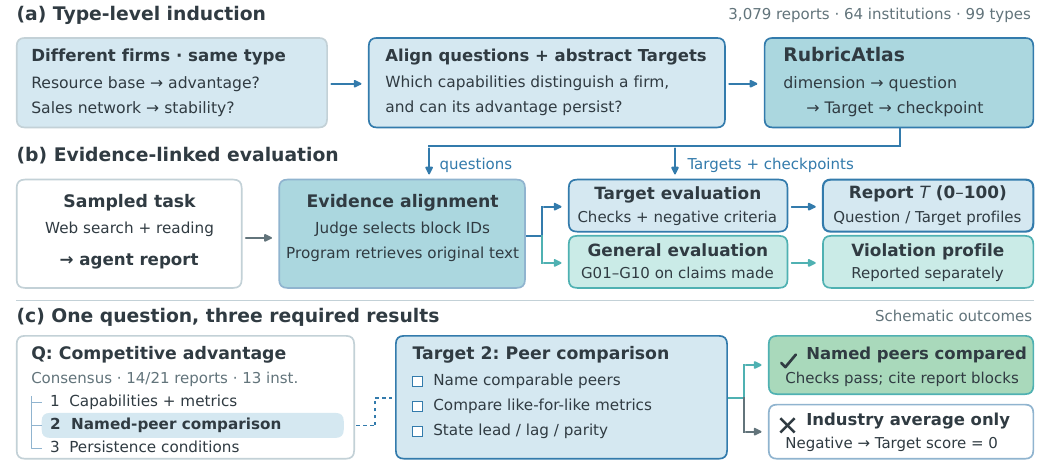}
\caption{Construction and evaluation. (a) Questions from expert reports on different firms are aligned and abstracted into type-level requirements. (b) Aligned report passages feed separate completeness and validity evaluations, yielding scores and analytical profiles. (c) An oil-and-gas question has three Targets; the peer-comparison Target is expanded into checkpoints. A named-peer comparison passes; an industry average triggers the negative criterion. Source questions are paraphrased; outcomes are schematic.}
\label{fig:frame}
\Description{Panel a shows questions about resource advantages and sales-network stability from different firms within one research type, aligned into a shared competitive-advantage question and abstracted into the RubricAtlas hierarchy. Panel b shows a sampled task producing a Web research report. A judge selects block IDs and a program retrieves original passages, which feed independent Target and General evaluations. Outputs are report completeness with question and Target profiles, and a separate violation profile. Panel c shows a consensus question supported by 14 of 21 reports from 13 institutions, with three Targets: capabilities and metrics, named-peer comparison, and persistence conditions. A dashed detail leader expands the second Target into three checkpoints. Schematic named-peer comparisons pass, while industry averages trigger a negative criterion that sets the Target score to zero.}
\end{figure*}

\begin{table*}[t]
\captionsetup{skip=2pt}
\caption{Source aggregation in related evaluations. All \atlas counts are medians.}
\label{tab:related}
\small
\renewcommand{\arraystretch}{0.95}
\setlength{\tabcolsep}{5pt}
\begin{tabular*}{\textwidth}{@{}l@{\extracolsep{\fill}}>{\raggedright\arraybackslash}p{0.61\textwidth}l@{}}
\toprule
Benchmark & Construction / reference basis & Scope\\
\midrule
DeepResearch Bench~\cite{du2025drbench} & LLM-generated, with one reference report & task\\
DRB II~\cite{li2026drbench2} & extracted from one expert report & task\\
ResearchRubrics~\cite{sharma2025researchrubrics} & expert-written for each prompt & prompt\\
FinResearchBench~\cite{sun2025finresearchbench} & report-level logic trees; shared structural metrics and LLM judgments & shared criteria\\
FinResearchBench~II~\cite{luan2026finresearchbench2} & multi-system reports for one query; consistency and distinguishability filters & query\\
\midrule
\atlas (ours) & \numReportsPerType{} expert reports per type; \supMedian{} supporting reports and \instMedian{} institutions per question & research type\\
\bottomrule
\end{tabular*}
\end{table*}

Our framework combines two designs (Figure~\ref{fig:frame}). \emph{Type-level induction} derives reusable analytical requirements across research objects. A \emph{question-centred hierarchy} organises their evaluation into questions, required results (\emph{Targets}) and binary checkpoints. For example, a question asks which capabilities give the firm an advantage; one required result is a peer comparison; its checkpoints ask whether the report names peers and provides comparable metrics. The details are thus evaluated as parts of an analysis, not as isolated facts. Passages are assigned to questions, checkpoint decisions cite the corresponding report blocks, and scores are aggregated within the hierarchy. This produces both a report score and a profile of completed and missing analyses. Completeness measures whether the required analytical content is present, not whether its facts or reasoning are correct. Argument validity is assessed separately.

We construct \atlas from \numReportsPool{} Chinese financial research reports issued by \numInstitutions{} institutions, yielding \numTypes{} research types, \numQuestions{} questions and \numItems{} checkpoints. Our reference-report selection experiments examine how the requirements used to evaluate these outputs depend on the expert reports behind them. We measure the fraction of a type's questions covered by each report and average within types; the median across types is \covReport\%. Evaluating only questions supported by each task's source report raises completeness scores by \inflSource{} points and gives Kendall $\tau$ = \tauSource{} with the full-atlas agent ranking. Thus, selecting one report's analyses changes both the coverage of the evaluation and its outcomes. These experiments restrict the frozen atlas; they do not reconstruct query-specific rubrics.

Our contributions are threefold:
\begin{itemize}
    \item \textbf{Type-level induction of analytical requirements.} We derive recurring analytical requirements across research objects and construct \atlas for financial research, with source provenance, tasks and evaluation artifacts.
    \item \textbf{Question-centred, evidence-linked report evaluation.} We organise detailed checks around analytical questions and retain question-level profiles alongside report scores. Two end-to-end judge pipelines achieve checkpoint agreement of $\kappa$ = \ckKappa{} and agent-ranking agreement of Kendall $\tau$ = \tauAB{}. Controlled deletions lower scores on targeted questions; padding produces no detectable increase. A first-author audit examines sampled source mappings and checkpoint decisions.
    \item \textbf{Reference sensitivity and Web-agent profiles.} We measure reference-selection effects on coverage, scores and rankings, and compare ten Web research agents on consensus and differentiating analyses. We examine associations with browsing, length and source-report exposure. Generic validity rules remain separate diagnostics because cross-judge agreement is limited.
\end{itemize}

\section{Related Work and Design Requirements}
\label{sec:desiderata}
Four requirements guide the framework: professional grounding, question-centred organisation and type-level reuse, attributable decisions, and reproducible comparisons. Table~\ref{tab:related} compares the source aggregation used by closely related evaluations.

\textbf{D1 Grounding.} A standard should represent recurring expert analyses, not one document's choices. Existing rubrics are written per instance~\cite{sharma2025researchrubrics,akyurek2025prbench}, extracted from one reference~\cite{li2026drbench2}, or mined from one survey per query~\cite{li2025researchqa}.

Multi-reference methods aggregate content for one input: Pyramid weights units by reference frequency~\cite{nenkova2004pyramid}, and TREC evaluates vital and okay nuggets~\cite{voorhees2003trecqa}. Nuggets can be weighted by assessor agreement~\cite{lin2006pyramids} or extracted and evaluated using LLMs~\cite{mayfield2024reports,pradeep2025nugget}. We abstract analyses across different objects within a type.

\textbf{D2 Organised and reusable.} Requirements should be organised around analytical questions, so that individual decisions reveal which analyses are complete or missing. This organisation should be reusable across tasks within a research type. Checklists turn holistic judgments into binary decisions~\cite{cook2024tick,lee2025checkeval,wei2025rocketeval}; instance-specific evaluators use rubrics or skill sets~\cite{ye2024flask,kim2024prometheus}, and rubrics also serve as rewards~\cite{gunjal2025rar,shao2025drtulu}. DEER and FinDeepResearch provide top-down task-level templates~\cite{han2025deer,zhu2025findeepresearch}. We retain binary checks but organise them under questions derived bottom-up from expert reports across research objects.

\textbf{D3 Attributable.} Judgments should identify the unmet requirement or supporting evidence. Factuality evaluation decomposes text into atomic claims~\cite{min2023factscore}; Mind2Web~2 verifies task-specific rubric trees against cited pages~\cite{gou2025mind2web2}, and ReportLogic evaluates reasoning~\cite{zhao2026reportlogic}. Our checkpoints cite report blocks retrieved verbatim by the program, not reconstructed by the judge.

\textbf{D4 Reproducible.} Comparisons should remain stable across judges. LLM judges exhibit position, verbosity and self-preference effects~\cite{zheng2023judging,wang2024fair,panickssery2024self,dubois2024lc} and criterion confusion~\cite{hu2024criteria}. IR studies examine assessor disagreement, incomplete judgments, topic sampling and metric discrimination~\cite{voorhees2000variations,zobel1998reliable,voorhees2002topic,sakai2006bootstrap}. Reproducibility differs from measurement validity~\cite{jacobs2021measurement}.

\textbf{Evaluation of Web research.} Web benchmarks assess short answers~\cite{wei2025browsecomp,mialon2023gaia} and deep research reports through rubrics or references~\cite{gou2025mind2web2,du2025drbench,li2026drbench2}. Financial benchmarks cover numerical question answering~\cite{chen2021finqa}, research reports~\cite{jin2026finrpt,zhu2025findeepresearch,jin2026finsight} and search~\cite{hu2025finsearchcomp}. We focus on the analytical completeness of reports produced through Web search and synthesis.

FinResearchBench~\cite{sun2025finresearchbench} combines report-level logic-tree metrics with LLM judgments. FinResearchBench~II~\cite{luan2026finresearchbench2} generates rubrics from multiple systems' reports for one query, then filters for judge consistency and distinguishability. Structured evaluation is therefore not unique to our approach: existing methods also use rubric trees, logic trees and templates. Our distinction is how the analytical structure is grounded and reused. Expert reports on different objects supply type-level questions and required results, under which evidence-linked checks are organised. The \emph{consensus} tier denotes analyses in a majority of source reports, not agreement among judges.

\section{Evaluation Framework}
\label{sec:frame}
The framework separates deriving an analytical standard from applying it. Type-level construction identifies recurring analyses across research objects; question-centred evaluation determines what each report delivers. Here, \emph{analytical structure} means the organisation of questions and required results, not the full logical coherence of a report's reasoning. Evaluation aligns report passages to this structure and judges completeness and argument validity independently.

\textbf{Research types.} A \emph{research-question family} is a recurring kind of analysis. Combining it with an industry defines a \emph{research type}, such as \emph{operations and value realisation of oil and gas producers}. A task specifies a firm, industry or policy within that type.

\textbf{Analytical requirements.} The rubric is a tree: \emph{dimension $\to$ question $\to$ required result $\to$ checkpoint}. Questions identify analyses, required results specify their expected outputs, and checkpoints assess the content needed to deliver those outputs. Each checkpoint decision therefore remains attached to an analytical requirement. A canonical \emph{question} represents an analysis found in the type's reports. Its \emph{support} records the reports and institutions addressing it; its \emph{tier} is \emph{consensus} if a majority of reports address it and \emph{differentiating} otherwise. A \emph{required result} (Target) specifies what an answer must deliver without naming a company or date. Binary \emph{checkpoints} state acceptance and non-acceptance conditions. A Target may have a \emph{negative criterion} identifying a substitute that misses the intended analysis, such as industry averages instead of named-peer comparisons. Requirements are constructed once per type, not per task.

\textbf{Evidence alignment and completeness.} The judge selects passage IDs; the program retrieves original text, preventing insertion of new material. Target evaluation checks content in these passages, citing blocks for \emph{yes} and explaining \emph{no}. Missed evidence during alignment can lower scores. Factual and inferential correctness are not assessed; Target requirements provide context, not extra scoring criteria.

In the current implementation, a Target's completeness is its checkpoint pass rate; a triggered negative criterion sets it to 0. Question completeness \T (0--1) is the mean of its Targets. Report completeness \T (0--100) is the weighted mean of question completeness (differentiating \wDiff, consensus \wCons) $\times 100$. Averaging within Targets and questions prevents extra checkpoints from increasing an analysis's report-level weight. Differentiating questions were model-screened and author-verified for analytical depth. The higher weight emphasises this depth and was fixed before examining agent evaluation results; alternatives appear in \S\ref{sec:bench}. Question, Target and checkpoint profiles locate incomplete analyses alongside report scores.

\textbf{Argument validity.} This layer concerns evidence, inference and quantitative reasoning. Flaw definitions can be shared, with question-specific conditions. \emph{General evaluation} tests generic rules on actual claims independently of Targets and scores (\S\ref{sec:flesh}).

\textbf{A worked example.} Figure~\ref{fig:frame} illustrates a consensus question for oil and gas producers: \emph{which verifiable capabilities give the firm an advantage over comparable firms, and can that advantage persist?} It is supported by \exSupport{} of \exPool{} reports from \exInst{} institutions, each on a different firm. Source questions concern a gas block's resource advantage or a sales network's contribution to stability. Its \exTargets{} Targets and \exItems{} checkpoints require capabilities with checkable metrics, named-peer comparisons on like-for-like metrics, and conditions sustaining or eroding the advantage. The second Target's negative criterion excludes industry averages in place of named peers. A report can thus receive credit for naming a capability but not for a peer comparison it never makes. Providing capability metrics does not substitute for comparing peers or examining persistence conditions: the separate Targets preserve these omissions within the same analytical question.

\textbf{Output.} We evaluate completeness end to end (\S\ref{sec:assessment}--\ref{sec:bench}); the separate validity profiles diagnose one generic-rule implementation (\S\ref{sec:flesh}).

\section{Constructing RubricAtlas for Financial Research}
\label{sec:atlas}
\atlas derives a shared analytical structure through report classification, question alignment and Target abstraction. Fixed AI workflows and prompts govern aggregation and applicability screening. Source questions determine the analyses to cover; answering passages supply each analysis's content requirements.

\textbf{Corpus and research types.} The corpus comprises \numReportsPool{} Chinese financial research reports on the Chinese market, issued by \numInstitutions{} institutions from January to September 2026. It covers \numIndustries{} industries and \numReportTypes{} genres, including company coverage, earnings reviews, industry deep dives and strategy reports.

An LLM decomposes reports into \emph{dimension $\to$ question $\to$ conclusion} trees. Reports are grouped first by genre $\times$ industry (\numCells{} cells), then by research-question family $\times$ industry. The \numQTypes{} families include firm operations and value realisation, industry cycle tracking and policy transmission. Of \numCandTypes{} candidates, \numDeferred{} with heterogeneous branches were deferred and \numDroppedTypes{} excluded by a predefined discrimination criterion. The \numTypes{} retained pools contain a median of \numReportsPerType{} reports (\poolMin--\poolMax; Appendix~\ref{app:types}).

\textbf{Question alignment.} Within each type, an LLM maps source questions to canonical questions, background or out-of-scope content; a second LLM checks the mappings. Records preserve supporting reports and source questions. Questions below \supThreshold{} supporting reports are removed unless judged to have distinct analytic value.

\textbf{Targets and checkpoints.} For each canonical question, answering passages are decomposed into \emph{data used $\to$ reasoning $\to$ conclusion}, abstracted into company- and date-independent Targets, and divided into checkpoints. We add negative criteria for plausible substitutes. Before scoring, an AI reviewer checked each criterion for one verifiable requirement, clear wording and no adjacent redundancy~\cite{akyurek2025prbench}; authors reviewed all flagged items. Scores use this frozen atlas, unchanged by the later author audit.

\textbf{Applicability screening.} The workflow excludes requirements outside the industry and research-question family, and those specific to an individual object. It also checks consistency between retained requirements and the requests' analytical scope. These filters produce type-level standards rather than report-content unions. Requirements are frozen before scoring; request checks match tasks to these standards without regenerating rubrics or adapting them to agent outputs.

\textbf{Scale and support.} \atlas contains \numTypes{} trees, \numQuestions{} questions (\numConsensus{} consensus, \numDiff{} differentiating), \numTargets{} Targets, \numItems{} checkpoints and \numForbidden{} negative criteria. Median question support is \supMedian{} reports and \instMedian{} institutions; \pctThreeInst\% of questions have at least three institutions and \pctFiveInst\% at least five. The largest institution's median support share is \maxInstShare\%.

Median pool coverage is \consReach\% for consensus and \diffReach\% for differentiating questions. The latter still have median support from 8 reports and \instDiff{} institutions (Table~\ref{tab:atlas}), representing recurring analyses rather than isolated contributions.

\begingroup
\setlength{\intextsep}{12pt}
\begin{table}[!ht]
\caption{Multi-source question support. Report and institution counts are medians per question; percentages concern differentiating questions.}
\label{tab:atlas}
\small
\begin{tabular*}{\columnwidth}{@{}l@{\extracolsep{\fill}}rrr@{}}
\toprule
Question tier & Questions & Reports & Institutions\\
\midrule
Consensus & \numConsensus & \supCons & 11\\
Differentiating & \numDiff & \textbf{8} & \textbf{\instDiff}\\
\midrule
\multicolumn{4}{@{}l@{}}{\emph{Multi-source support among differentiating questions}}\\
\multicolumn{3}{@{}l}{At least three supporting reports} & \textbf{\diffGeThree\%}\\
\multicolumn{3}{@{}l}{At least three supporting institutions} & \textbf{\diffThreeInst\%}\\
\bottomrule
\end{tabular*}
\end{table}
\endgroup

\textbf{Tasks.} A seeded hash selects one report per type and converts its topic into a conclusion-free request, yielding \numTasks{} tasks. Checks cover research type, object scope, analytical scope and readability, with recorded adjustments (Appendix~\ref{app:types}). Requests are matched to the type-level requirements, not used to regenerate them.

\textbf{Resource contents and access.} The resource includes trees with provenance, rules, tasks, prompts, checkpoint judgments, scoring and analysis code\ifpending{} at \url{\repoURL}\fi, and a datasheet~\cite{gebru2021datasheets}. The materials will be made publicly available after the review process. Re-scoring uses judgments and code; independent re-judging and source verification also need report text and source materials. A verbatim-overlap check sampled \verbN{} criteria, extracted \verbSpans{} spans of 20 normalised characters and searched \verbTexts{} source texts: no matches, with a positive control detected.

\section{Experimental Setup}
\label{sec:setup}
\textbf{Agents and harness.} The ten LLMs in Table~\ref{tab:main} run as deep research agents in a fixed ReAct harness~\cite{yao2023react}. \emph{WebSearch} returns Google results via Serper; \emph{WebFetch} reads pages with Jina and extracts goal-relevant content with \texttt{gpt-5}. The prompt sets information cutoff \cutoffDate, requires citable sources and prohibits other institutions' research conclusions as evidence. Each agent runs once per task with provider-default decoding, up to \maxTurns{} turns and \maxTokens{} output tokens per turn. Tasks and reports are Chinese.

\textbf{Evaluation sets.} One run failed, leaving \numReports{} reports. Agent and browsing comparisons use \numCommon{} common tasks (980 reports), covering \dropSourceQTotal{} of \numQuestions{} atlas questions. Checkpoint agreement uses available paired judgments. Agent scores equally average common-task report scores; tier scores are averaged after within-report computation, so global tier counts cannot reconstruct totals.

\textbf{Judges and protocol.} Judge~A is DeepSeek-V4.1-Flash; B is GPT-6-luna. Calls use temperature 0 and are grouped by dimension, with decisions assigned to questions. \emph{Evidence alignment} selects block IDs for programmatic text retrieval. Target evaluation returns checkpoint and negative-criterion decisions, citations for each \emph{yes} and reasons for each \emph{no}. General evaluation runs separately. Neither judge accesses the Web.

The API parameter \texttt{reasoning\_effort} is \emph{none} for alignment, \emph{medium/low} for A's Target/General calls and \emph{high/high} for B's. Model-specific labels are not a shared budget scale: we compare configured pipelines, not matched-budget model substitution. Settings were fixed in pilots before finalisation; token use and alternatives appear in Appendix~\ref{app:reruns}. Failed jobs are listed separately, not scored zero.

A is also an evaluated backbone; B shares the tree builders' GPT family. Both judges expose sensitivity to these overlaps but cannot rule out family bias. All agents share the \texttt{gpt-5} page extractor, whose effect is not isolated. Task resampling quantifies sampling uncertainty; score intervals add a shared repeat-run variance proxy (\S\ref{sec:judges}), not independently estimated B variance.

\section{Assessing the Resource and Evaluation Protocol}
\label{sec:assessment}
We assess two aspects of the approach: which analytical requirements the resource represents, and how consistently evaluation applies them. Reference-report selection analyses examine requirement coverage and dependence on particular expert reports; the author audit examines sampled mappings and decisions; cross-judge comparisons and controlled content changes assess the evaluation protocol.

\subsection{Author audit}
\label{sec:authoraudit}
The first author, a finance graduate student, annotated blank forms without AI assistance or access to model judgments: 300 checkpoint decisions and 297 source mappings for 50 canonical questions. Agreement with A/B is 74.3\%/71.7\% ($\kappa$ = 0.49/0.43). Mappings are 77.8\% faithful, 15.2\% partial and 7.1\% not faithful. These checks concern checkpoint application and question grouping, not every Target abstraction or task applicability.

The sample overrepresents judge disagreements; evidence is the union of both judges' aligned passages. Agreement cannot be directly compared with the full-corpus judge--judge rate. All 300 labels are retained as final judgments. This domain-informed author audit is not independent expert validation; sampling, confusion matrices and exploratory preferences appear in Appendix~\ref{app:authoraudit}.

\subsection{Reference-report selection and analytical coverage}
\label{sec:onereport}
How does reference selection change requirement coverage and agent comparisons? Using the full atlas as an internal reference, we retain questions supported by selected reports and re-score frozen main-run judgments (\S\ref{sec:setup}). Reports and checkpoint decisions stay fixed; rubrics are not re-extracted (Appendix~\ref{app:procedures}). These comparisons measure reference sensitivity, not which requirements best fit each task.

\textbf{Coverage.} Averaging reports within types, then taking the median across types gives \covReport\% coverage (type means \covTypeMin--\covTypeMax\%); task-source reports cover \covSource\%. The gap is not solely expert disagreement: genre medians are \covGenreHi\% for company coverage and \covGenreLo\% for retrospective reviews; the shortest and longest thirds cover \covShort\% and \covLong\%.

Removing the measured report from support and dropping questions below \supThreshold{} reports leaves median coverage at \covLOTO\%. The other half covers \covHalfOut\% of questions selected from one random half of a pool; the within-half threshold retains more common questions. A single report thus represents only part of a type's recurring analyses, even excluding its contribution to question selection.

\textbf{Score shifts.} Source-supported subsets raise mean completeness by \inflSource{} points (task-level 95\% confidence interval (CI) \ci{\inflSourceCIlow}{\inflSourceCIhigh}), with increases on \inflPosTasks\% of tasks. Size-matched random subsets give \inflRandom{} (\numDraws{} draws; maximum \inflRandomMax). Source subsets contain \consShareSrc\% consensus questions, versus \consShareAll\% overall; consensus scores are higher. Composition matching gives \compInfl{} (maximum \compInflMax{} across \numDraws{} draws), still \inflBeyondComp{} below source subsets, which exceed every matched draw. The remainder is consistent with tasks eliciting source-emphasised analyses, but object-specific applicability is not isolated.

\textbf{Ranking sensitivity.} Single-report subsets have lower ranking agreement and more leader changes than size-matched random questions (Table~\ref{tab:kcurve}). Task-source subsets give \tauSource{}, versus \tauRandomSubset{} for size-matched subsets and mean \compTau{} (5th percentile \compTauP) for composition-matched subsets. Size matching alone does not isolate source dependence: composition explains part of the ranking change. Source subsets change the task-level leader on \srcTopOneChanged\% of tasks, versus \randTopOneChanged\% for size-matched random subsets (Appendix~\ref{app:pertype}). Both reductions affect task-level recommendations, motivating aggregate comparisons.

\begin{table}[t]
\caption{$k$-report question subsets per type (\numDraws{} draws; smaller pools use all reports). Cov.: mean union coverage over types/draws, not median per-report coverage. $\tau$: mean correlation with the full ranking; 5th: fifth percentile. Rand.: count-matched random questions, without tier matching. Gain: mean score increase (points). Top-1: fraction changing the overall leader.}
\label{tab:kcurve}
\small
\setlength{\tabcolsep}{5pt}
\begin{tabular}{@{}rrrrrrrr@{}}
\toprule
$k$ & Cov. & $\tau$ & $\tau$ (5th) & $\tau$ rand. & Gain & Gain rand. & Top-1 (rand.)\\
\midrule
1 & \kOneCov\% & \kOneTau & \kOneTauP & \kOneTauRand & \kOneInfl & \kOneInflRand & 24\% (7\%)\\
3 & \kThreeCov\% & \kThreeTau & \kThreeTauP & \kThreeTauRand & \kThreeInfl & \kThreeInflRand & 0\% (0\%)\\
10 & \kTenCov\% & \kTenTau & \kTenTauP & \kTenTauRand & \kTenInfl & \kTenInflRand & 0\% (0\%)\\
\bottomrule
\end{tabular}
\end{table}

\begin{figure}[t]
\includegraphics[width=\columnwidth]{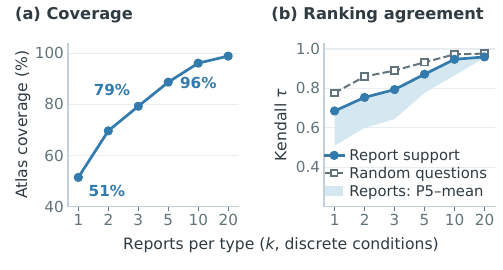}
\caption{Full-atlas coverage by $k$ reports per type (a) and ranking Kendall $\tau$ (b), with size-matched random questions. Shading spans the report draws' 5th percentile to mean. Equally spaced ticks denote discrete $k$ conditions; smaller pools use all reports.}
\label{fig:kcurve}
\Description{Two line plots show atlas coverage and ranking correlation as the number of supporting reports per type increases. The ranking plot includes a size-matched random question-subset comparison and a shaded band down to the fifth percentile of report-support draws.}
\end{figure}

\textbf{Aggregation across reports.} Three and ten reports cover \kThreeCov\% and \kTenCov\% of the atlas (Figure~\ref{fig:kcurve}), with ranking correlations \kThreeTau{} and \kTenTau{}, near size-matched controls (\kThreeTauRand{}, \kTenTauRand). Single-report small pools (\poolMin--\smallPoolHi) give \ttau{} = \smallPoolTau{} and \smallPoolCov\% coverage, but sensitivity is non-monotonic (\midPoolLo--\midPoolHi: \midPoolTau; \largePoolLo--\largePoolHi: \largePoolTau). About ten reports recover much of the full pool's coverage and ranking: an internal planning reference, not an external sufficiency threshold.

\textbf{Sensitivity to source reports.} Disjoint random half-pools give \ttau{} = \tauHalf{} across \numDraws{} splits (\tauHalfThree{} with three supporting reports within each half), always with the same leader. Task resampling gives \tauTaskResample{} against the full ranking, a similar scale of variation.

Removing task-source support drops \dropSourceQ{} of \dropSourceQTotal{} questions across \numCommon{} common tasks below threshold; re-scoring gives \ttau{} = \tauDropSource{}, with model score changes at most \maxScoreChangeDropSource{} points. The aggregate evaluation is therefore not driven by the single report used to construct each task. This supports type-level standards; new-task applicability is discussed in \S\ref{sec:scope}.

\subsection{Cross-judge consistency and repeat runs}
\label{sec:judges}
B independently re-evaluates all \numReports{} reports, including alignment. The \numCkCompared{} paired checkpoints agree at \ckAgree\%, with Cohen's $\kappa$ = \ckKappa{} (95\% CI \ci{\ckKappaLo}{\ckKappaHi}, report-clustered) and Jaccard overlap \ckJaccard{} for \emph{yes} decisions. Question-level \T correlation is \qCorrT{} over \numQCompared{} questions.

B gives fewer \emph{yes} decisions (\ckYesB\% versus \ckYesA\%) and scores reports \levelGap{} points lower on average, but rankings agree at \ttau{} = \tauAB{} (95\% CI \ci{\tauABlo}{\tauABhi}, task resamples). Both judges have the same first two agents, top-four set and bottom-three order (Table~\ref{tab:main}, Figure~\ref{fig:judges}a). Of A's \pairsOne{} pairs separated by over one point, \flipsOne{} reverses under B; none of \pairsTwo{} separated by over two points reverses. None of B's \pairsOneB{} pairs separated by over one point reverses under A. Judge choice shifts absolute levels more than clearly separated agents' ordering. Negative criteria rarely trigger (A \ncTrigA, B \ncTrigB{} of \numNcCompared); agreement is \ncAgree\% ($\kappa$ = \ncKappa).

A DeepSeek repeat run on \jnN{} stratified reports (\jnTasks{} tasks $\times$ ten agents) uses the same model and effort but condensed negative-criterion wording, correlating at \jnCorr{} with the main run. Its standard deviation (SD) proxy is \jnSD{} points per report and \jnModelSD{} for a \numCommon-task mean. Both judges use this estimate, mixing run and wording variation. Alternative settings correlate highly, but nominal high effort need not increase tokens (Appendix~\ref{app:reruns}).

\subsection{Response to deletion and padding}
\label{sec:perturb}
We perturb \ptN{} reports (ten tasks $\times$ three agents) and re-evaluate each whole report. Deleting all blocks answering one question lowers its \T from \delBeforeA{} to \delAfterA{} under A and from \delBeforeB{} to \delAfterB{} under B. Median relative reductions are \delRelA\% and \delRelB\%, with decreases on \delKA/\delNA{} and \delKB/\delNB{} target questions\footnote{Perturbation targets were drawn before the atlas was frozen; \delNA{} of the 30 targeted questions are in the resource atlas and are analysed here.}. Questions sharing no deleted block change by \delOtherA{} and \delOtherB{} on average, close to unchanged re-evaluation (\delOtherNoopA{} and \delOtherNoopB). The response is concentrated on questions whose content was removed, linking the score change to an identifiable missing analysis rather than only a reduction in report length; Appendix~\ref{app:reruns} also reports changes on questions sharing deleted blocks.

\begin{figure*}[t]
\includegraphics[width=\textwidth]{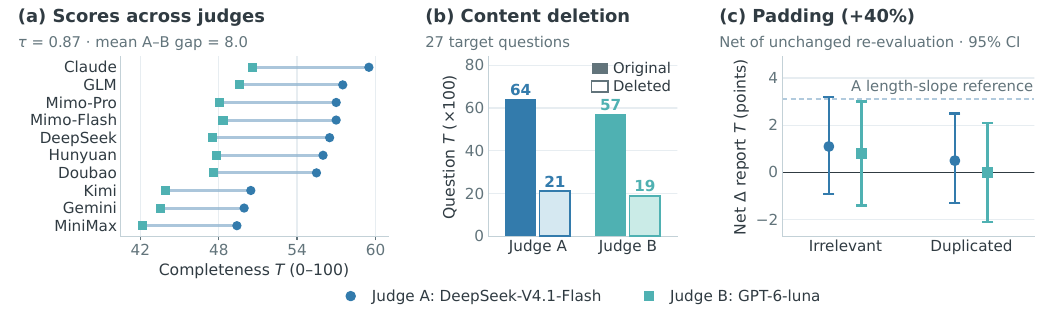}
\caption{Completeness diagnostics (levels on 0--100). (a) Ten agents' report \T, ordered by A; Kendall $\tau$ summarises ranking agreement. (b) Mean completeness on \delNA{} target questions before (solid) and after (pale) deletion. (c) Net report \T changes after \ptPad\% irrelevant or duplicated text, with paired 95\% intervals against unchanged re-evaluation. The dashed A length-slope reference is not an equivalence margin.}
\label{fig:judges}
\Description{A paired dot plot displays the same ten agents' completeness scores under judges A and B. A grouped bar plot compares mean target-question completeness before and after deletion. Four padding point estimates have paired confidence intervals, a zero line, and a dashed reference calculated from judge A's observed length slope.}
\end{figure*}

Adding \ptPad\% irrelevant text from another research family or duplicating paragraphs gives no detectable net increase. A's unchanged re-evaluation itself shifts scores by \noopA{} points. Relative to each judge's unchanged re-evaluation, irrelevant/duplicated effects are \padIrrNetA{} \ci{\padIrrNetLoA}{\padIrrNetHiA} and \padDupNetA{} \ci{\padDupNetLoA}{\padDupNetHiA} under A, \padIrrNetB{} \ci{\padIrrNetLoB}{\padIrrNetHiB} and \padDupNetB{} \ci{\padDupNetLoB}{\padDupNetHiB} under B (Figure~\ref{fig:judges}c). With \ptN{} reports, only effects above about 3 points are excluded. A's observed length slope implies \padExpectA{} points, so modest rewards remain compatible with the intervals.

\section{Evaluating Deep Research Agents}
\label{sec:bench}
We compare the final reports of Web research agents through completeness totals and analytical tier profiles. We then examine associations with browsing, report length and exposure to source reports. Generic-rule profiles remain separate from completeness.

\subsection{Main results}
\begin{table*}[t]
\caption{Completeness \T and tier scores on \numCommon{} common tasks (0--100). $\pm$: 95\% half-width from task bootstrap (\numBoot{} resamples) and shared repeat-run variance. Rank: task-resampling 95\% range; P(1st): first-place fraction. Cons./Diff.: consensus/differentiating scores. Rows follow unrounded judge-A means. Absolute levels depend on the judge.}
\label{tab:main}
\small
\setlength{\tabcolsep}{4pt}\begin{tabular}{@{}lrrcrrrrrcr@{}}
\toprule
& \multicolumn{6}{c}{Judge A (DeepSeek-V4.1-Flash)} & \multicolumn{4}{c}{Judge B (GPT-6-luna)}\\
\cmidrule(lr){2-7}\cmidrule(l){8-11}
Agent & \T & $\pm$ & Rank & P(1st) & Cons. & Diff. & \T & $\pm$ & Rank & P(1st)\\
\midrule
Claude-Opus-5.5 & \scClaude & \hwClaude & \rkloClaude--\rkhiClaude & \pfClaude\% & \ctClaude & \dtClaude & \luClaude & \luhwClaude & \lurkloClaude--\lurkhiClaude & \lupfClaude\%\\
GLM-5.3 & \scGLM & \hwGLM & \rkloGLM--\rkhiGLM & \pfGLM\% & \ctGLM & \dtGLM & \luGLM & \luhwGLM & \lurkloGLM--\lurkhiGLM & \lupfGLM\%\\
Mimo-V2.6-Pro & \scMimoPro & \hwMimoPro & \rkloMimoPro--\rkhiMimoPro & \pfMimoPro\% & \ctMimoPro & \dtMimoPro & \luMimoPro & \luhwMimoPro & \lurkloMimoPro--\lurkhiMimoPro & \lupfMimoPro\%\\
Mimo-V2.6-Flash & \scMimoFlash & \hwMimoFlash & \rkloMimoFlash--\rkhiMimoFlash & \pfMimoFlash\% & \ctMimoFlash & \dtMimoFlash & \luMimoFlash & \luhwMimoFlash & \lurkloMimoFlash--\lurkhiMimoFlash & \lupfMimoFlash\%\\
DeepSeek-V4.1-Flash & \scDeepSeek & \hwDeepSeek & \rkloDeepSeek--\rkhiDeepSeek & \pfDeepSeek\% & \ctDeepSeek & \dtDeepSeek & \luDeepSeek & \luhwDeepSeek & \lurkloDeepSeek--\lurkhiDeepSeek & \lupfDeepSeek\%\\
Hunyuan-4-Preview & \scHunyuan & \hwHunyuan & \rkloHunyuan--\rkhiHunyuan & \pfHunyuan\% & \ctHunyuan & \dtHunyuan & \luHunyuan & \luhwHunyuan & \lurkloHunyuan--\lurkhiHunyuan & \lupfHunyuan\%\\
Doubao-Seed-2.1-Pro & \scDoubao & \hwDoubao & \rkloDoubao--\rkhiDoubao & \pfDoubao\% & \ctDoubao & \dtDoubao & \luDoubao & \luhwDoubao & \lurkloDoubao--\lurkhiDoubao & \lupfDoubao\%\\
Kimi-K3 & \scKimi & \hwKimi & \rkloKimi--\rkhiKimi & \pfKimi\% & \ctKimi & \dtKimi & \luKimi & \luhwKimi & \lurkloKimi--\lurkhiKimi & \lupfKimi\%\\
Gemini-3.8-Flash & \scGemini & \hwGemini & \rkloGemini--\rkhiGemini & \pfGemini\% & \ctGemini & \dtGemini & \luGemini & \luhwGemini & \lurkloGemini--\lurkhiGemini & \lupfGemini\%\\
MiniMax-M3 & \scMiniMax & \hwMiniMax & \rkloMiniMax--\rkhiMiniMax & \pfMiniMax\% & \ctMiniMax & \dtMiniMax & \luMiniMax & \luhwMiniMax & \lurkloMiniMax--\lurkhiMiniMax & \lupfMiniMax\%\\
\bottomrule
\end{tabular}
\end{table*}

\textbf{Leading agents and unresolved margins.} Claude leads under both judges (\pfClaude\%, \lupfClaude\% first-place resamples), followed by GLM. Their gaps of \marginA{} and \marginB{} points remain unresolved: 95\% intervals \ci{\marginAlo}{\marginAhi}, \ci{\marginBlo}{\marginBhi} include shared repeat-run variance. Ranks 3--7 span \spreadThreeSevenA{} and \spreadThreeSevenB{} points, separated from Kimi, Gemini and MiniMax at ranks 8--10 by \gapSevenEightA{} and \gapSevenEightB{} points under A/B.

Claude remains first under consensus/differentiating weights 1:1, 1:3 and 2:1, inverse-frequency and frequency weighting (\ttau{} \tauWeightTwoOne--\tauWeightOneOne). Frequency weighting emphasises recurring source analyses (\tauWeightFreq); consensus-only scoring puts GLM first (\tauConsOnly; Appendix~\ref{app:robust}). The evaluated analyses affect rankings even with a stable leader.

\textbf{Distinct analytical profiles.} Similar consensus scores can conceal differences on differentiating analyses. Under A, GLM, Doubao and Claude are within half a point on consensus questions, leading in \consFirstShareGLMA\%, \consFirstShareDoubaoA\% and \consFirstShareClaudeA\% of resamples. Claude leads differentiating questions by \diffLeadA{} points under A (\diffFirstShareA\% first-place resamples) and \diffLeadB{} under B (\diffFirstShareB\%).

Differentiating scores have greater dispersion across agents: \diffSDA{} versus \consSDA{} points under A (difference \sdGapA{}, 95\% CI \ci{\sdGapLoA}{\sdGapHiA}) and \diffSDB{} versus \consSDB{} under B (\sdGapB{}, \ci{\sdGapLoB}{\sdGapHiB}). Dispersion is the population SD across evaluated agents. B's interval reaches zero, so the stronger statistical evidence comes from A. Reporting both tiers shows which groups of analyses separate agents with similar totals. These profiles summarise coverage of recurring requirements, not the coherence or validity of the agents' reasoning.

\subsection{Browsing, length and completeness}
Claude makes \searchesClaude{} searches and \fetchesClaude{} page reads per report; Gemini makes \searchesGemini{} and \fetchesGemini. Portals and financial media supply \shPortal\% of reads, official sources \shOfficial\%. Within tasks, log searches and reads correlate with \T at \corrSearchT{} and \corrFetchT{} under A, \corrSearchTB{} and \corrFetchTB{} under B. Official-source share is weakly associated ($r$ = \corrOfficialT).

With agent/task fixed effects and task-clustered standard errors (SEs), A's page-read coefficient is \regFetchA{} points per log unit (SE \regFetchASE), versus \regSearchA{} for searches (SE \regSearchASE, not significant). Adding log length gives page coefficient \regFetchLenA{} (SE \regFetchLenASE) and length coefficient \regLenA{} (SE \regLenASE). B's page coefficient is \regFetchB{} (SE \regFetchBSE, not significant). Browsing evidence is stronger under A; page reads and length correlate at $r$ = \corrFetchLen{} within tasks; these observational data do not isolate their contributions.

Longer reports score higher: \lenCoefA{} points per log length unit under A (SE \lenSEA), \lenCoefB{} under B (SE \lenSEB). This association does not identify a causal length effect; padding is tested in \S\ref{sec:perturb}. Length adjustment leaves \lenLeaderA{} first under both judges, with Mimo-V2.6-Pro second under A and \lcSecondB{} under B (\ttau{} = \lcTauA{}, \lcTauB{} with main rankings). Length-matched win rates are \matchedClaude\% for Claude and \matchedGLM\% for GLM (Appendix~\ref{app:robust}); Claude's lead does not carry over to matching, although it remains first after regression adjustment.

\textbf{Judge differences and self-preference.} Between-judge score shifts are associated with length. DeepSeek's length-adjusted shift does not establish self-preference; Appendix~\ref{app:robust} reports estimates and intervals.

\subsection{Do agents retrieve the source reports?}
Research-report sites account for \leakPageShare\% of \numFetches{} successful reads and occur in \leakReportsAny\% of generated reports' trajectories. Sell-side content appears in \leakSellShare\% of reports, with scores \leakSellDelta{} points above task means, consistent with browsing associations. Task-source titles appear in search results or pages in \leakSrcN{} of \numTraj{} trajectories (\leakSourceTitle\%, 95\% CI \ci{\leakSrcLo}{\leakSrcHi}); exposed reports show no detectable gain (\leakSrcEff{} points, \ci{\leakSrcEffLo}{\leakSrcEffHi}). Source-title exposure is uncommon with no detected benefit. The prompt prohibits other institutions' conclusions, but its effect is not isolated.

\subsection{Generic validity-rule diagnostics}
\label{sec:flesh}
Completeness records delivered analyses, not their validity. General evaluation applies ten argument-flaw rules covering facts, evidence and reasoning (G01--G10; Appendix~\ref{app:rules}) to aligned material in separate calls (A low, B high effort). It reports claim blocks, basis blocks and reasons per question, without Target requirements, scores or treating missing content as a violation.

\textbf{Limited agreement on individual rules.} Cross-judge $\kappa$ has median \ruleKappaMed{} and maximum \ruleKappaMax{} (G04, calculations); flagged-question Jaccard has median \ruleJacMed{} (Table~\ref{tab:rules}). A and B flag \ruleHitsDS{} and \ruleHitsLuna{} question--rule violations. For G09 (source qualifications), they flag \rhNineA{} and \rhNineB{} questions, with \rhNineBoth{} shared. Despite shared definitions, rule decisions are less consistent than completeness checkpoints ($\kappa$ = \ckKappa) on the same reports.

\textbf{Agent-level violation profiles.} To assess whether aggregation reduces disagreement, we compare each agent's fraction of questions with any violation, separately from completeness. A flags \flagMinA--\flagMaxA\%, versus B's \flagMinB--\flagMaxB\%; rates depend strongly on the judge. Flag-rate rankings give Kendall \ttau{} = \flagTau{} (95\% CI \ci{\flagTauLo}{\flagTauHi}; $r$ = \flagR), versus completeness \tauAB{} \ci{\tauABlo}{\tauABhi}. The intervals overlap, so the point-estimate difference does not establish a ranking-level contrast; the clearer difference is in individual checkpoint and rule agreement ($\kappa$ \ckKappa{} versus median \ruleKappaMed). Across rules, agent-level $\tau$ has median \mtMed{} and range \mtNine{} (G09) to \mtThree{} (G03; Table~\ref{tab:rules}). These are diagnostics, not validity scores: combining them with completeness would introduce judge dependence without validated weights.

\begin{table}[t]
\caption{Generic rules on \numQCompared{} shared questions: A/B flags, overlap and Cohen's $\kappa$. Agent level: flag-rate Kendall $\tau$ and Pearson $r$ (ten agents).}
\label{tab:rules}
\small
\setlength{\tabcolsep}{4.5pt}
\begin{tabular}{@{}lrrrrrr@{}}
\toprule
& \multicolumn{4}{c}{Question level} & \multicolumn{2}{c}{Agent level}\\
\cmidrule(lr){2-5}\cmidrule(l){6-7}
Rule & A & B & Both & $\kappa$ & $\tau$ & $r$\\
\midrule
G04 & \rhFourA & \rhFourB & \rhFourBoth & \rhFourKap & \mtFour & \mrFour\\
G06 & \rhSixA & \rhSixB & \rhSixBoth & \rhSixKap & \mtSix & \mrSix\\
G02 & \rhTwoA & \rhTwoB & \rhTwoBoth & \rhTwoKap & \mtTwo & \mrTwo\\
G07 & \rhSevenA & \rhSevenB & \rhSevenBoth & \rhSevenKap & \mtSeven & \mrSeven\\
G03 & \rhThreeA & \rhThreeB & \rhThreeBoth & \rhThreeKap & \mtThree & \mrThree\\
G10 & \rhTenA & \rhTenB & \rhTenBoth & \rhTenKap & \mtTen & \mrTen\\
G01 & \rhOneA & \rhOneB & \rhOneBoth & \rhOneKap & \mtOne & \mrOne\\
G05 & \rhFiveA & \rhFiveB & \rhFiveBoth & \rhFiveKap & \mtFive & \mrFive\\
G08 & \rhEightA & \rhEightB & \rhEightBoth & \rhEightKap & \mtEight & \mrEight\\
G09 & \rhNineA & \rhNineB & \rhNineBoth & \rhNineKap & \mtNine & \mrNine\\
\bottomrule
\end{tabular}
\end{table}

\textbf{Descriptive diagnostics.} G07, G03, G08 and G04 give relatively consistent agent profiles, unlike G09 and G01. Both judges flag \flagLeastA{} least often. These describe detected flaws, not validated question judgments (Appendix~\ref{app:rules}).

\textbf{Interpreting the gap.} Checkpoints specify conditions; generic rules leave thresholds such as ``overstated'' to judges. Binary presence checks differ from multi-label defect detection. Checkpoints pass about half the time; most rules flag fewer than one question in six. A uses medium Target and low General effort, B high for both. These differences are plausible explanations, not separately tested causes.

\section{Discussion and Limitations}
\label{sec:scope}
\textbf{Implications for evaluation.} Web research reports need evaluation beyond retrieved facts. Type-level requirements and question-centred checks locate incomplete analyses, preserving profiles behind report scores. These diagnose output content, not whether omissions originate in retrieval or synthesis. Tier profiles locate agent differences on sampled requests. Keeping validity separate prevents conflating missing content with defective reasoning. Absolute scores should identify the judge; relative comparisons are most credible where judges agree.

\textbf{Scope and transfer.} The resource covers Chinese-language financial research (\numIndustries{} industries, \numQTypes{} question families). Applicability screening and request-scope checks are part of construction, but their accuracy has not been independently quantified. Sampled requests do not establish applicability to every new object. New types need report pools and construction. About ten reports is an internal planning reference, not a transfer guarantee (Table~\ref{tab:threats}).

\textbf{Beyond financial research.} A broader direction is type-level rubric induction as a reusable paradigm for document evaluation. Academic papers, literature surveys~\cite{li2025researchqa}, legal documents and other expert-written texts could supply genre-specific requirements, with completeness traced to evidence and argument validity assessed separately. Adapting this paradigm beyond finance will require domain-specific corpora, applicability checks and validation of both layers.

\textbf{Mining versus task-generated rubrics.} An LLM could instead generate a rubric for each task, including a structured rubric. Our distinction is not the presence of a hierarchy alone, but the basis for its analytical questions: model-generated expectations for one request versus analyses documented across expert reports on different objects. Mining preserves source-question records for auditing each abstraction, while the question-centred hierarchy connects detailed checks to those analyses. This is a design argument, not a head-to-head comparison.

\textbf{Developing validity evaluation.} Question-specific checks, such as whether growth follows from stated inputs, require separate validation. Once both layers are validated, \emph{Pareto dominance} compares agents no worse on either dimension and strictly better on at least one, otherwise retaining the trade-off. Future directions include crediting checkpoints only when cited passages have no established flaw, and propagating unsupported premises through question dependencies.

\textbf{Limitations.} The first-author audit is not independent expert validation and does not establish task applicability or the fidelity of every Target. Source support measures recurrence, not importance; tier weights are a design choice. Source experiments restrict a frozen atlas rather than re-extract single-report or task-generated rubrics. We do not isolate the contribution of hierarchical organisation against a flat-checklist baseline. Perturbations test unrelated and duplicated text, not on-topic empty expansion. Rule effort and output format are not isolated by ablation. Shared repeat-run variance is a proxy that includes wording changes. Agents run once per task, so intervals exclude generation randomness; search has no date filter.

\textbf{Ethics and access.} Materials will be made publicly available after the review process. Trees omit analyst names and use opaque institution IDs. Completeness does not assess investment suitability. The audit involved only the first author, no external annotators.

\section{Conclusion}
Expert reports on different objects supply recurring analytical requirements, not merely reference answers. \atlas organises these requirements by type and links checks to analytical questions, yielding profiles of completed and missing analyses in Web deep research reports. Reference-report selection experiments motivate cross-report aggregation; cross-judge evaluation and controlled deletions support evidence-linked completeness comparisons. A first-author audit adds a human check on sampled mappings and decisions. Argument validity remains separate: the tested generic rules require further validation before entering a combined score.

\clearpage
\bibliographystyle{ACM-Reference-Format}
\bibliography{references}

\appendix
\setlength{\intextsep}{10pt plus 2pt minus 2pt}
\setlength{\textfloatsep}{12pt plus 2pt minus 2pt}
\setlength{\floatsep}{10pt plus 2pt minus 2pt}
\captionsetup[table]{skip=4pt}
\section{Generic Argument-Flaw Rules}
\label{app:rules}
\begin{center}
\small
\setlength{\tabcolsep}{4pt}
\begin{tabular}{@{}l>{\raggedright\arraybackslash}p{0.87\columnwidth}@{}}
\toprule
Rule & What is checked\\
\midrule
G01 & Factual and citation accuracy\\
G02 & Consistency of objects, metrics and baselines\\
G03 & Realisation status, timing and conditions\\
G04 & Checkability of key calculations\\
G05 & Necessary inferential steps\\
G06 & Overstated causal and contribution claims\\
G07 & Support for conclusion scope and certainty\\
G08 & Treatment of material conflicts and uncertainties\\
G09 & Source qualifications for the claims made\\
G10 & Information validity at the research cutoff\\
\bottomrule
\end{tabular}
\end{center}
The judge sees report material without Web access. G01 covers errors identifiable from the text or common knowledge; G10 checks time references against the cutoff.

\textbf{Descriptive diagnostics.} Agent-level $(\tau,r)$ is (\mtSeven, \mrSeven) for G07, (\mtThree, \mrThree) for G03, (\mtEight, \mrEight) for G08 and (\mtFour, \mrFour) for G04. B flags G08 on \mrateBEightPct\% of questions; G04 rates are \mrateAFour\%/\mrateBFour\%. Both judges flag Doubao least (\flagDoubaoA\%/\flagDoubaoB\%). Question-level $\kappa\le$\ruleKappaMax. These descriptive profiles are more consistent for report-internal features; G09/G01 can require external knowledge. They do not independently validate the rules.

\section{Aggregation Robustness}
\label{app:robust}
Table~\ref{tab:tierweights} contrasts reweighting both tiers with retaining one tier, using A's fixed judgments.

\begin{table}[!ht]
\caption{Aggregation sensitivity under A. Ratios are consensus:differentiating; $\tau$ compares with the main agent ranking.}
\label{tab:tierweights}
\small
\setlength{\tabcolsep}{4pt}
\begin{tabular*}{\columnwidth}{@{}l@{\extracolsep{\fill}}cc l@{}}
\toprule
Scheme & Cons.:Diff. & $\tau$ & First\\
\midrule
\multicolumn{4}{@{}l}{\emph{Both tiers retained}}\\
Equal & 1:1 & \tauWeightOneOne & Claude\\
More differentiating & 1:3 & \tauWeightOneThree & Claude\\
More consensus & 2:1 & \tauWeightTwoOne & Claude\\
Inverse frequency & varies & \tauWeightInvFreq & Claude\\
Frequency & varies & \tauWeightFreq & Claude\\
\midrule
\multicolumn{4}{@{}l}{\emph{One tier retained}}\\
Consensus only & 1:0 & \tauConsOnly & GLM\\
Differentiating only & 0:1 & \tauDiffOnly & Claude\\
\bottomrule
\end{tabular*}
\end{table}

For \matchedPairs{} same-task pairs within \matchedRatio\% in length, win rates are \matchedGLM\% for GLM, \matchedDoubao\% for Doubao and \matchedClaude\% for Claude; their order remains unresolved. Log-length regression with agent/task fixed effects retains Claude first, followed by Mimo-V2.6-Pro under A ($\tau$ = \lcTauA{} with main ranking) and \lcSecondB{} under B (\lcTauB). Matching and regression adjustment are distinct, not a single corrected ranking.

Agents' A-minus-B scores correlate with mean log report length ($r$ = \shiftLenR). Relative shifts are \shiftClaude{} \ci{\shiftloClaude}{\shifthiClaude} for Claude, the longest writer, \shiftKimi{} \ci{\shiftloKimi}{\shifthiKimi} for Kimi, the shortest, and \shiftDeepSeek{} \ci{\shiftloDeepSeek}{\shifthiDeepSeek} for DeepSeek. DeepSeek's length-adjusted shift is \dsResidual{} \ci{\dsResidualLo}{\dsResidualHi}: no detected self-preference beyond length, but up to two points remains compatible.

\textbf{Group averaging.} The current scoring manifest places each checkpoint in its own group. Averaging within groups and then across groups therefore equals the checkpoint pass rate for each Target before the negative-criterion gate.

\section{Judge Prompt Structure}
\label{app:prompts}
\textbf{Evidence alignment.} The archive prompt assigns report blocks to questions within a rubric dimension, returning IDs without evaluation.

\textbf{Checkpoint evaluation.} Decisions use full acceptance conditions, alternatives and exclusions. The judge cannot supply missing content from domain knowledge. It cites block IDs for \emph{yes}, gives reasons for \emph{no} and checks the negative criterion separately.

\textbf{General evaluation.} The prompt checks errors in actual claims, without Targets or scores. Findings identify question, rule, claim blocks, basis blocks and reason. Resource prompts produce JSON outputs subject to programmatic validation.

\section{Re-runs, Perturbations and Execution Failures}
\label{app:reruns}
\textbf{Repeat-run protocol.} With a fixed seed, we sample ten research-question families without replacement, one task per family rotating across pool-size tertiles, and all ten agents (\jnN{} reports). Mean main-run \T is \jnLevel{}, below the benchmark average; re-evaluation lowers scores by \jnShift{} points, with a shared agent shift. Perturbations use three agents on these ten tasks.

Repeat-run and perturbation experiments retain the main-run checkpoints but use condensed negative-criterion wording.

\textbf{Variance proxy.} Dividing the SD of paired repeat-minus-main report scores by the square root of two gives a report-level proxy of \jnSD{} points. Dividing this by the square root of \numCommon{} gives \jnModelSD{} points for the mean over these tasks. These steps assume independent, equal-variance run errors and independent errors across tasks. Score-interval half-widths combine the task-bootstrap half-width and 1.96 times the mean proxy by summing their squares and taking the square root. Agent-gap intervals additionally scale the mean proxy by the square root of two, assuming independent agent errors. Ranks and first-place fractions use task resampling alone. Both judges share A's proxy, which includes wording variation; B-specific, correlated and generation errors are not separately estimated.

\textbf{Perturbation results.} Deletion lowers target-question \T on \delKA/\delNA{} questions under A (95\% Wilson interval~\cite{wilson1927probable} \delLoA--\delHiA\%) and \delKB/\delNB{} under B (\delLoB--\delHiB\%). Questions sharing deleted blocks also lose credit: \delShareAQ{} and \delShareBQ{} on average, versus \delShareNoopA{} and \delShareNoopB{} under unchanged re-evaluation.

Padding changes from the original run are \padIrrA{} \ci{\padIrrLoA}{\padIrrHiA} (irrelevant) and \padDupA{} \ci{\padDupLoA}{\padDupHiA} (duplicated) under A, and \padIrrB{} \ci{\padIrrLoB}{\padIrrHiB} and \padDupB{} \ci{\padDupLoB}{\padDupHiB} under B. A's unchanged re-evaluation shifts scores by \noopA{}. Figure~\ref{fig:judges}c reports paired net effects against each judge's unchanged re-evaluation.

\textbf{Reasoning settings and failures.} High effort uses \tokHighAexact{} tokens under A versus \tokMedAexact{} at medium (Table~\ref{tab:reasoning}). B's high setting uses fewer tokens than A's medium setting; labels do not imply matched budgets.

\begin{table}[!ht]
\caption{Settings on \jnN{} reports. Tokens: mean reasoning tokens per Target call; $r$: report-\T correlation with A's main run on the paired sample.}
\label{tab:reasoning}
\small
\setlength{\tabcolsep}{4pt}
\begin{tabular*}{\columnwidth}{@{}l@{\extracolsep{\fill}}lrr@{}}
\toprule
Judge & Target/General effort & Tokens & $r$\\
\midrule
A & low/none & \tokLowA & \rbLowA\\
A & medium/low & \tokMedA & \jnCorr\\
A & high/high & 8.1k & \rbHighA\\
B & medium/low & \tokMedB & \rbMedB\\
B & high/high & \tokHighB & \rbFullB\\
\bottomrule
\end{tabular*}
\end{table}

Output-validation failures (\rejB{} in the judge-B run, \rejNoise{} in the repeat run and \rejPert{} in perturbations) are listed separately, not assigned zero.

\section{Research Types and Task Sampling}
\label{app:types}
The \numTypes{} types combine \numQTypes{} research-question families with \numIndustries{} industries. Pools contain a median of \numReportsPerType{} reports (range \poolMin--\poolMax). The type list records pool size, question count, tier composition, median support and institutions per question.

\textbf{Standardised screening.} Fixed AI workflows and prompts check industry/question-family relevance, cross-object applicability and request scope (\S\ref{sec:atlas}), without regenerating task rubrics or adapting standards to agent outputs.

Reports are ordered within each type by the SHA-256 hash of the seed, category ID and report ID, concatenated with vertical-bar separators. The seed is \nolinkurl{dr-bench-99-tasks-20261009-v1}. The sampling record documents research-type, object-scope and readability checks, including three object adjustments and six wording changes on 2026-10-10. The incomplete task belongs to a type with nine questions, leaving \dropSourceQTotal{} questions across the \numCommon{} common types.

\section{Further Discussion}
\label{app:discussion}
\begin{table}[t]
\caption{Main threats to validity, the evidence we report and the residual risk.}
\label{tab:threats}
\small
\begin{tabular}{@{}>{\raggedright\arraybackslash}p{0.2\columnwidth}>{\raggedright\arraybackslash}p{0.42\columnwidth}>{\raggedright\arraybackslash}p{0.3\columnwidth}@{}}
\toprule
Threat & Evidence & Residual risk\\
\midrule
Circular coverage & Held-out \covLOTO\%; out-of-half \covHalfOut\%; composition-matched control & Split halves reuse the atlas questions\\
Judge dependence & Two families: $\kappa$ = \ckKappa, $\tau$ = \tauAB; re-run SD \jnSD{} points & Author audit is not independent validation\\
Length & Padding estimate \padIrrNetA{} points; length-adjusted leader unchanged & Modest reward not excluded; matched and adjusted rankings differ\\
Self-preference & DeepSeek shift net of length \dsResidual{} \ci{\dsResidualLo}{\dsResidualHi} & One agent is also a judge\\
Leakage & \leakPageShare\% of reads; no detected source-title exposure gain & Search API has no date filter\\
Frequency $\neq$ importance & Model-screened, author-verified depth; weights fixed before agent results; same leader under five weightings & Weights remain a design choice\\
Scope & Fixed screening; matched request scope across \numIndustries{} industries & One domain/language; screening unvalidated\\
\bottomrule
\end{tabular}
\end{table}

\begin{figure}[t]
\centering
\includegraphics[width=\columnwidth]{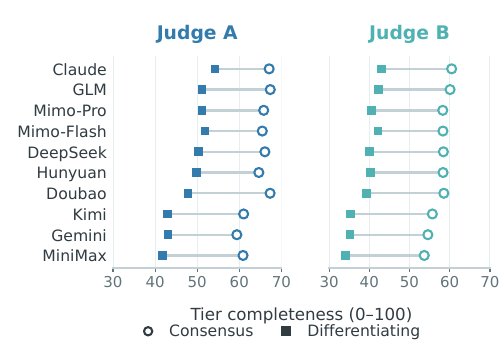}
\caption{Tier completeness under both judges on \numCommon{} common tasks. Paired means are computed within reports, then averaged over tasks. Panels share a score axis and follow Table~\ref{tab:main}'s judge-A ordering.}
\label{fig:tiers}
\Description{Two horizontal paired-dot panels show ten agents under judges A and B. Open circles indicate consensus completeness and filled squares indicate differentiating completeness. Grey lines pair the two tier means for each agent. Both panels use the same agent order and an axis from 30 to 70 on the 0 to 100 score scale. The figure displays descriptive means without confidence intervals or significance markers.}
\end{figure}

\textbf{Interpreting profiles.} Figure~\ref{fig:tiers} shows both tiers. Under A, GLM and Doubao are within half a point of Claude on consensus questions, but trail by about three and six on differentiating questions. B's overall scores are \levelGap{} points lower. Requirement reduction affects task-level recommendations (Appendix~\ref{app:pertype}); comparisons aggregate \numCommon{} tasks.

\textbf{Nugget evaluation.} TREC nuggets define content for one input~\cite{voorhees2003trecqa,mayfield2024reports,pradeep2025nugget}; our questions abstract analyses across objects. Assessor agreement~\cite{lin2006pyramids} and report support measure recurrence, but scoring uses tier weights (\S\ref{sec:frame}), not institution counts. A comparison can satisfy required metrics yet use inappropriate peers: its completeness and validity are distinct.

\section{Split-Half, Leave-One-Out, Source-Removal and \texorpdfstring{$k$}{k}-Report Procedures}
\label{app:procedures}
Table~\ref{tab:sourceprotocols} summarises reference selection with frozen candidate questions and judgments. No rubrics are re-mined or reports re-judged.

\begin{table}[!ht]
\caption{Reference-report procedures. Fixed: original tiers retained. Split-half recomputes tiers from within-half support.}
\label{tab:sourceprotocols}
\small
\setlength{\tabcolsep}{3pt}
\begin{tabular}{@{}>{\raggedright\arraybackslash}p{0.23\columnwidth}>{\raggedright\arraybackslash}p{0.55\columnwidth}>{\raggedright\arraybackslash}p{0.16\columnwidth}@{}}
\toprule
Procedure & Questions retained & Tiers\\
\midrule
Leave-one-out & Remove the measured report's support; drop questions below \supThreshold{} reports & Not used\\
Split-half & Support from at least two reports in the half (three in the variant) & Within-half\\
Source removal & Remove task-source support; drop questions below \supThreshold{} reports & Fixed\\
$k$ reports & Supported by at least one of $k$ randomly sampled reports & Fixed\\
Task source & Supported by the report used to construct the task & Fixed\\
\bottomrule
\end{tabular}
\end{table}

Leave-one-out measures held-out report coverage. Split-half recomputes tiers using majority support, falling back to median support when tier counts differ by more than half the retained questions and the fallback gives a nonempty consensus tier.

Sampling $k$ reports is without replacement (all reports for smaller pools). Controls draw equally many random questions per task. Task-source subsets use the report supplying the task (Appendix~\ref{app:pertype}).

\section{Single-Reference Rubrics and Per-Type Recommendations}
\label{app:pertype}
Restricting evaluation to questions supported by the task's source report changes the task-level leader on \srcTopOneChanged\% of tasks and the top-three set on \srcTopThreeChanged\%. Size-matched random subsets change the leader on \randTopOneChanged\% of tasks (5th--95th percentiles \randTopOneLo--\randTopOneHi\%). Task-level rankings are therefore sensitive to both tested reductions of the requirements, motivating our focus on aggregate agent comparisons.

\section{First-Author Audit: Protocol and Results}
\label{app:authoraudit}
The first author is a graduate student in finance and also participated in designing the framework. Annotation began from blank forms, without AI assistance or access to the model judgments. This is an author audit blind to model labels, not independent expert validation. It was conducted after the evaluated resource was fixed and does not alter the atlas, model judgments or agent scores.

\textbf{Checkpoint sampling and materials.} The 300 items were sampled with seed 20261023 from atlas checkpoints judged by both models. Quotas were 100 both-yes, 100 both-no, 50 A-yes/B-no and 50 A-no/B-yes. Each report--question contributed at most one item and each task at most six. Questions with no aligned passage under either judge were excluded. The author received the union of the passages aligned by both judges, not necessarily the same material either judge had evaluated. Disagreement can therefore reflect evidence selection as well as interpretation of a checkpoint.

\textbf{Final labels.} The author returned 177 yes and 123 no labels, with no abstentions or blanks. All 300 labels are confirmed as final judgments. Agreement is 223/300 with A and 215/300 with B, with $\kappa$ = 0.487 and 0.433. Table~\ref{tab:authoraudit} gives the confusion counts and stratified agreement. The model judges agree on 200/300 items (66.7\%) in this deliberately disagreement-enriched sample, rather than the full-corpus \ckAgree\%.

\begin{table}[!ht]
\caption{First-author checkpoint audit. Confusion counts use the author's label as the row and model label as the column, not as an adjudicated ground truth. The lower panel reports agreement within sampling strata.}
\label{tab:authoraudit}
\small
\begin{tabular*}{\columnwidth}{@{}l@{\extracolsep{\fill}}rr@{}}
\toprule
Author / model labels & Judge A & Judge B\\
\midrule
Yes / yes & 125 & 121\\
Yes / no & 52 & 56\\
No / yes & 25 & 29\\
No / no & 98 & 94\\
\midrule
Both judges yes ($n=100$) & 91\% & 91\%\\
Both judges no ($n=100$) & 78\% & 78\%\\
A yes, B no ($n=50$) & 68\% & 32\%\\
A no, B yes ($n=50$) & 40\% & 60\%\\
\bottomrule
\end{tabular*}
\end{table}

Of the items, 166 concern consensus questions and 134 differentiating questions. Author--A agreement is 74.1\% and 74.6\%, respectively ($\kappa$ = 0.468/0.500); author--B agreement is 68.7\% and 75.4\% ($\kappa$ = 0.359/0.514). All items are included in the primary result.

\textbf{Reweighting.} The stored checkpoint population contains 36,457 both-yes, 33,149 both-no, 8,814 A-yes/B-no and 3,049 A-no/B-yes items. Weighting the four sample agreement rates by these proportions gives 81.3\% for A and 78.2\% for B. These auxiliary estimates do not correct the report--question and task caps or the exclusion of empty evidence, and are not unbiased estimates of population accuracy.

\textbf{Source-question mappings.} The audit sampled 25 consensus and 25 differentiating questions, at most one per research type and six source mappings per question. Among 297 mappings, 231 were marked yes (77.8\%), 45 partial (15.2\%) and 21 no (7.1\%). Partial mappings are reported separately: a source question may address only part of a broader canonical question. This audit examines question grouping, not every Target's fidelity to source arguments or applicability to a task object.

\textbf{Report preferences.} The optional audit included 20 same-task report pairs stratified by judge-A score gaps below 3, from 3 to 8, and at least 8 points. The author's final judgments preferred the first report in seven pairs, the second in six, and judged seven comparable. Among 13 directional preferences, eight (61.5\%) matched A's completeness ordering; the seven ties are not counted as errors.

This small author-only sample does not validate a general expert-preference ranking.

\end{document}